%% file: main.tex
\documentclass[11pt]{article}

\usepackage[table]{xcolor}
\usepackage[preprint]{acl}
\usepackage{times}
\usepackage{latexsym}
\usepackage[T1]{fontenc}
\usepackage[utf8]{inputenc}
\usepackage{microtype}
\usepackage{inconsolata}
\usepackage{graphicx}
\usepackage{amsmath}
\usepackage{amssymb}
\usepackage{booktabs}
\usepackage{tabularx}
\usepackage{multirow}
\usepackage{float}
\usepackage{adjustbox}
\input{tables/table_style.tex}
\usepackage{pifont}

\DeclareRobustCommand{\gearr}{\texorpdfstring{\ensuremath{\mathrm{GEAR}^{2}}}{GEAR2}}

\newcommand{\gear}{GEAR}

\newif\ifhasreferences
\hasreferencestrue

\title{GEAR\textsuperscript{2}: Pruned Experts Live On through Gated Reuse of Activated Experts and a Missing-Mass Residual}
\author{%
  Linliang Chen\textsuperscript{1,*},
  Yan Zhong\textsuperscript{2,*\textdagger},
  Wang Kang\textsuperscript{3,\textdaggerdbl},
  Sai Li\textsuperscript{3,\textdaggerdbl},
  Ji Wu\textsuperscript{4},
  Ji Liu\textsuperscript{4} \\
  \textnormal{\textsuperscript{1}Hangzhou International Innovation Institute, Beihang University, Hangzhou, China} \\[-2pt]
  \textnormal{\textsuperscript{2}School of Mathematical Sciences, Peking University, Beijing, China} \\[-2pt]
  \textnormal{\textsuperscript{3}School of Integrated Circuit Science and Engineering, Beihang University, Beijing, China} \\[-2pt]
  \textnormal{\textsuperscript{4}Department of Electronic Engineering, Tsinghua University, Beijing, China} \\[-3pt]
  \textnormal{\small \textsuperscript{*}Equal contribution.\quad
    \textsuperscript{\textdagger}Project lead.\quad
    \textsuperscript{\textdaggerdbl}Corresponding authors.} \\[-2pt]
  \texttt{\small\mdseries linliang.chen@buaa.edu.cn, zhongyan@stu.pku.edu.cn, \{wkang,saili\}@buaa.edu.cn} \\[-5.5pt]
  \texttt{\small\mdseries wuji\_ee@mail.tsinghua.edu.cn, jiliuwork@gmail.com}%
}

\begin{document}
\maketitle

\begin{abstract}
Mixture-of-Experts (MoE) Large Language Models (LLMs) activate only a few experts per token while generally keeping all the experts in GPU memory for high efficiency. Expert pruning reduces this GPU memory footprint cost, yet discards the contribution of pruned experts. Existing remedies either rely on static substitutes, merge experts at the cost of altering their functions, or require costly recovery training. In this paper, we find that the functionalities of pruned experts can be efficiently reconstructed from the activated remaining experts. We propose \gear{} (\underline{G}ated \underline{E}xpert \underline{A}ctivation \underline{R}euse), which reuses the scaled outputs of reliable activated experts without any training or additional expert computation. In addition, we propose \gearr{} to further recover what reuse cannot express with an ultra-lightweight residual. On five MoE LLMs from 16B to 106B parameters with 25\% and 50\% of experts pruned, GEAR improves over its training-free counterparts in nearly every setting (up to 4~pp.) in terms of accuracy, and \gearr{} outperforms baseline approaches (up to 2.75~pp.) in every setting on nine zero-shot benchmarks, with larger advantages (up to 2.72~pp. higher) under more aggressive pruning (up to 50\%).
\end{abstract}

\section{Introduction}
\label{sec:introduction}

Mixture-of-Experts (MoE) Large Language Models (LLMs) activate only a few experts for each token, trading bounded computation for vast parameter capacity, and MoE has become the mainstream architecture of LLMs~\citep{gptoss,mixtral,deepseekv2,qwen3,glm45}. 
However, this design comes with two notable challenges: (i)~MoE trades memory for FLOPs, so all experts are generally loaded to GPU memory at deployment for efficiency, which incurs a large GPU memory footprint and makes decoding bound by the memory traffic of expert weights~\citep{mone}; (ii)~expert usage is imbalanced, leaving substantial redundancy across experts~\citep{hcsmoe}. Expert pruning, which removes redundant experts entirely, is the most direct remedy for both~\citep{naee,reap}, but it discards the contribution of pruned experts to the tokens routed to them. In this work, we focus on compensating for the pruned experts under a limited budget, so that pruned MoE LLMs retain the memory savings and inference speedup of pruning while approaching the capabilities of the original models.

\begin{figure*}[t]
  \centering
  \includegraphics[width=0.8\textwidth]{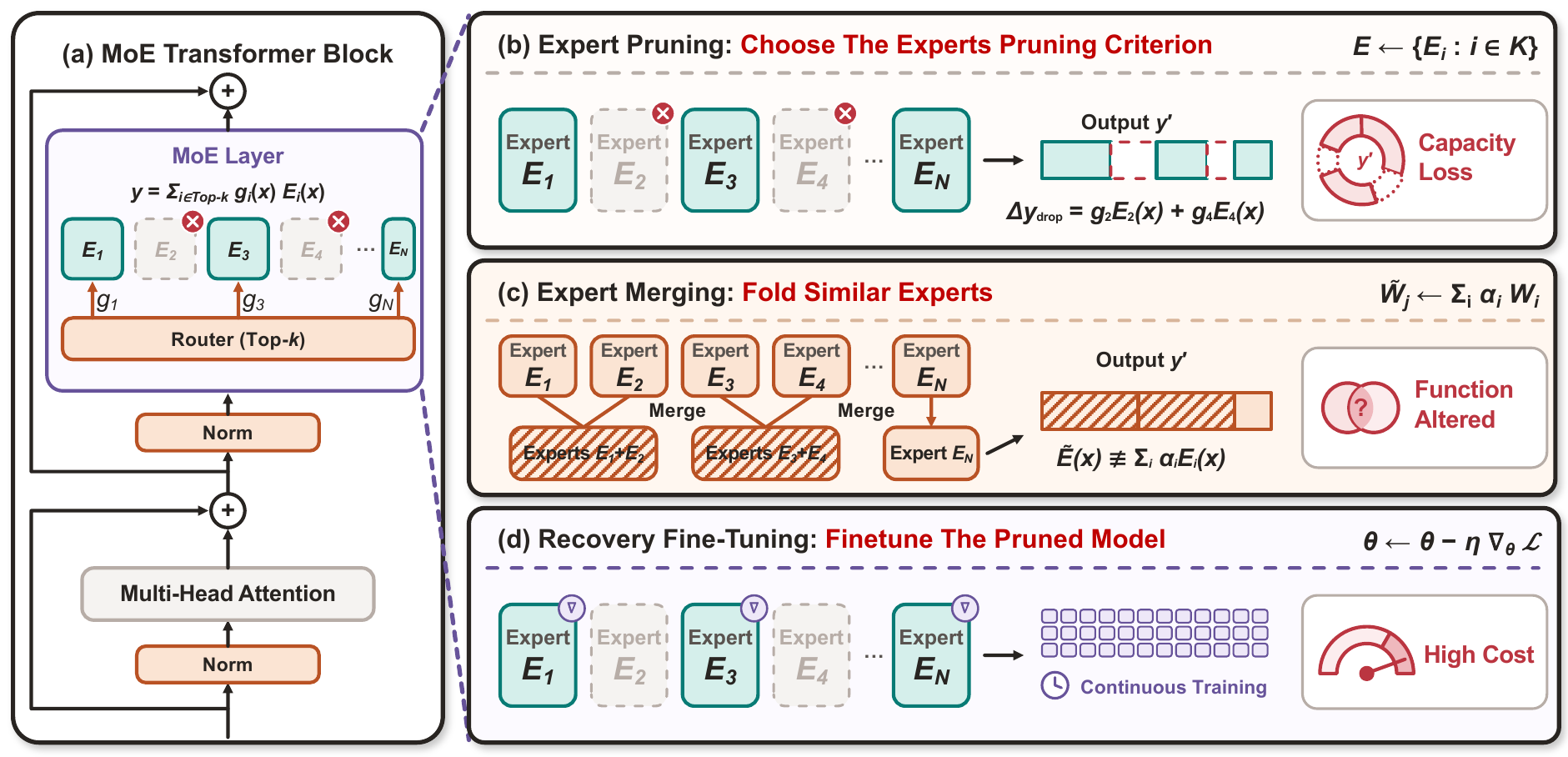}
  \caption{Existing routes for mitigating the loss from expert pruning. (a)~An MoE Transformer block. (b)~Pruning discards the contribution of pruned experts. (c)~Merging alters the functions of merged experts. (d)~Recovery fine-tuning is costly.}
  \label{fig:motivation}
\end{figure*}

Existing work mitigates the performance degradation from pruning along three routes (Figure~\ref{fig:motivation}b--d). First, pruning methods design better criteria for which experts to remove~\citep{naee,reap,fantastic}, and some other approaches replace each pruned expert with a substitute, such as its mean output or an affine transform of a retained expert~\citep{mone,mora}. These substitutes are static, i.e., they do not adapt to the experts activated for the current token, so the functionalities of pruned experts are only partly restored, and the performance degradation grows with the pruning ratio. Second, merging approaches fold functionally similar experts into one, so that their knowledge survives in the merged weights~\citep{mcsmoe,hcsmoe}. However, since experts are nonlinear functions, averaged parameters cannot reproduce the weighted sum of the original outputs~\citep{reap}. In addition, the difference between the original experts and the newly generated merged experts propagates across layers~\citep{routerkd}. Third, recovery training regains most of the performance~\citep{fantastic,slimqwen}, but it updates all parameters with substantial costs.

Our key hypothesis is that the functionality of a pruned expert does not vanish with its parameters: experts in an MoE LLM form groups of functionally similar experts~\citep{hcsmoe,diversify,stun}. The functionality of a pruned expert can be covered by other experts in the same Top-$k$ combination~\citep{shape}. Thus, it can be largely reconstructed from the experts activated for the same token.

We thus propose a training-free Gated Expert Activation Reuse (\gear{}) approach and its lightweight-training extension \gearr{}, which dynamically compensate for pruned experts (Figure~\ref{fig:method}). To recover the functionalities of the pruned experts, we decompose the functionalities of a pruned expert into a \emph{shared component} and a \emph{non-collinear residual}. The shared component represents the functionalities of pruned experts that exist in other retained experts. The non-collinear residual represents the other functionalities of pruned experts that are hard to recover from the retained experts. To reconstruct the shared component of pruned experts, \gear{} performs \emph{expert reuse}, i.e., the shared functionalities of each pruned expert are largely recovered by reusing the scaled output of its \emph{best proxy} among the activated experts. A \emph{reliability gate} falls back to the mean output when the proxy is unreliable. The gate is essential as unreliable proxies make reuse worse than no reuse at all. 
Then, in order to recover the non-collinear residual, we propose \gearr{} to exploit an ultra-lightweight compensation module trained by a small amount of end-to-end distillation.

Our contributions are summarized as follows:
{\setlength{\leftmargini}{1.1em}\begin{itemize}\setlength{\itemsep}{0pt}\setlength{\parskip}{0pt}\setlength{\topsep}{2pt}
  \item We propose \gear{}, which compensates for pruned experts by reusing their best proxies among the activated experts behind a reliability gate, with no training or extra computation.
  \item We propose \gearr{}, i.e., an ultra-lightweight compensation module, which recovers the non-collinear residual by missing-mass weighting and light distillation.
  \item We conduct extensive experiments with five MoE models ranging from 16B to 106B and nine zero-shot tasks. Experimental results demonstrate superior performance of \gear{} compared with its training-free counterparts in nearly every setting, and \gearr{} consistently outperforms four baseline approaches, with larger advantages at higher pruning ratios.
\end{itemize}}

\section{Related Work}
\label{sec:related-work}

\paragraph{Expert pruning.}
Weight-level pruning removes individual weights within experts~\citep{sparsegpt,wanda,moepruner}, but the resulting irregular sparsity is hard to translate into actual speedup. Expert pruning instead removes entire experts and directly reduces storage. Existing methods differ mainly in how they select the experts to remove, using routing statistics~\citep{nllb,holistic}, the impact on layer outputs~\citep{naee,stun,reap}, or collaboration among experts~\citep{shape}. Some methods further replace each pruned expert with a mean output~\citep{mone} or an affine transform of a fixed proxy expert~\citep{mora}. Such substitutes are fixed offline and do not adapt to the experts actually activated for the current token.

\paragraph{Expert merging.}
Merging methods fold functionally similar experts into one, so that the knowledge of pruned experts is retained in the merged parameters~\citep{mcsmoe,hcsmoe,submoe,dern}. This line of work also reveals that experts form groups with similar functions~\citep{hcsmoe,diversify}. However, merging rewrites the retained experts and alters their own functions. Since experts are nonlinear, merged parameters cannot reproduce the input-dependent mix of the original outputs, and this error causes the functional subspace of later layers to collapse~\citep{reap}; the merged result is also fixed offline.

\paragraph{Recovery training after compression.}
Recovery training continues to train the compressed model~\citep{fantastic,lightmoe,slimqwen}, or updates only a small set of parameters such as the router~\citep{routerkd} or low-rank adapters~\citep{llmpruner,recoverlora}. However, continued training approaches the cost of retraining, often with hundreds of billions of tokens~\citep{slimqwen}, consuming the very cost that pruning was meant to save. Router calibration only re-allocates the remaining experts and does not restore the outputs of pruned ones. Low-rank adapters apply their correction regardless of how much routing weight a token loses, giving no stronger correction to tokens that lose more.

\paragraph{Dynamic expert skipping at inference.}
Another line of work keeps all experts and reduces those executed per token, by using a variable number of experts~\citep{topp,adamoe,moepp}, skipping low-weight routes~\citep{naee}, or folding routes onto executed experts~\citep{lynx,sere,exfold}. These methods do not reduce memory; after pruning, the contribution of removed experts can only be recovered through compensation.

\section{Preliminaries}
\label{sec:preliminaries}

\paragraph{MoE layer.}
An MoE layer~\citep{moe,switch} consists of $N$ routed experts $\{E_1,\dots,E_N\}$, a router, and optionally shared experts. Given a layer input $x\in\mathbb{R}^d$, the router scores every expert, selects the $k$ highest-scoring ones as the Top-$k$ set $\mathcal{T}(x)$, and assigns routing weights $g_i(x)$. The layer output is the weighted sum of the selected experts' outputs:
\begin{equation}
  y(x)=y_s(x)+\sum_{i\in\mathcal{T}(x)}g_i(x)\,E_i(x),
  \label{eq:moe}
\end{equation}
where $y_s(x)$ is the output of the shared experts (zero if absent).

\paragraph{Expert pruning.}
Expert pruning keeps a subset of experts $\mathcal{K}\subset\{1,\dots,N\}$ in each layer according to some criterion and removes the rest. We keep the router unchanged, so each token still selects $\mathcal{T}(x)$ by its original scores. This set splits into the activated retained experts $\mathcal{R}(x)=\mathcal{T}(x)\cap\mathcal{K}$ and the pruned experts $\mathcal{D}(x)=\mathcal{T}(x)\setminus\mathcal{K}$, and Eq.~\eqref{eq:moe} becomes
\begin{equation}
  \begin{aligned}
  y(x)={}&y_s(x)+\sum_{j\in\mathcal{R}(x)}g_j(x)\,E_j(x)\\
  &+\underbrace{\sum_{i\in\mathcal{D}(x)}g_i(x)\,E_i(x)}_{c(x)}.
  \end{aligned}
  \label{eq:pruned}
\end{equation}
The first two terms are computed as before, while the last term, the \emph{contribution of the pruned experts} $c(x)$, disappears with their parameters.

\paragraph{Compensation objective.}
Given a pruning mask, we compensate for $c(x)$ under three constraints: (i)~the retained experts and the router stay unchanged, so compensation applies to any mask; (ii)~the original routing weights $g_i(x)$ are kept without renormalization, so the router still weights the compensation of each pruned expert; and (iii)~compensation executes no additional experts at inference and uses little calibration and training data.

\section{Method}
\label{sec:method}

\begin{figure*}[t]
  \centering
  \includegraphics[width=0.85\textwidth]{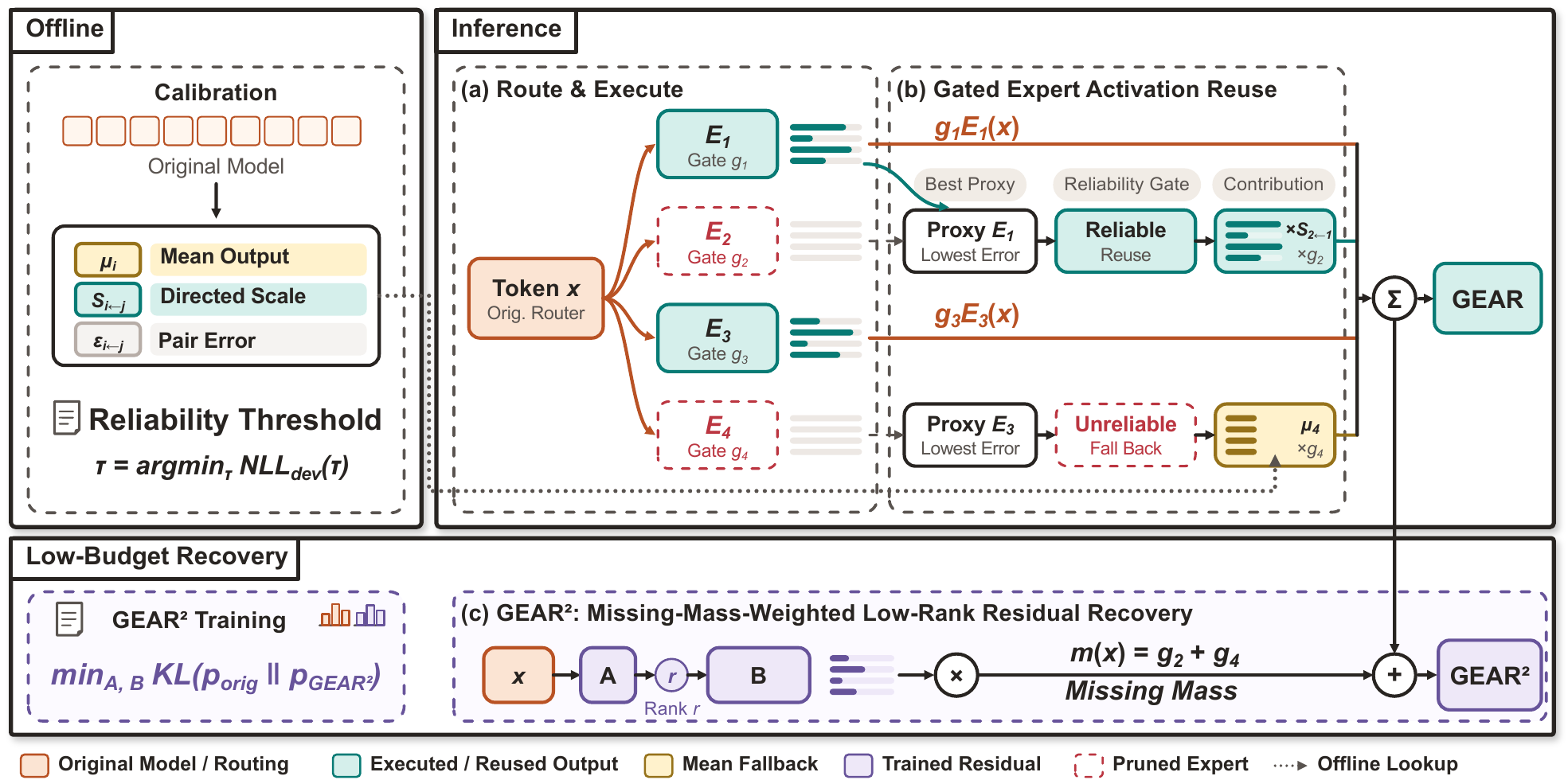}
  \caption{Overview of GEAR and \gearr{}. (a)~The pruned model keeps its router and activated retained experts; (b)~each pruned expert reuses the scaled output of a reliable proxy and falls back to the mean otherwise; (c)~\gearr{} adds a missing-mass-weighted low-rank residual, trained by light distillation with all original parameters frozen.}
  \label{fig:method}
\end{figure*}

Figure~\ref{fig:method} illustrates the pipeline. We keep the output of the pruned model, $y(x)-c(x)$, and replace the lost contribution $c(x)$ with an estimate $\hat{c}(x)$, built from an estimate $\hat{E}_i(x)$ of each pruned expert and a residual term $\Delta(x)$:
\begin{equation}
  \begin{aligned}
  \hat{y}(x)&=y(x)-c(x)+\hat{c}(x),\\
  \hat{c}(x)&=\sum_{i\in\mathcal{D}(x)}g_i(x)\,\hat{E}_i(x)+\Delta(x).
  \end{aligned}
  \label{eq:framework}
\end{equation}
GEAR and \gearr{} provide the two terms of $\hat{c}(x)$. GEAR (\S\ref{sec:gear}) constructs $\hat{E}_i(x)$ by reusing the outputs of the experts activated for the current token; it needs only one offline calibration and no training, with $\Delta(x)=0$. \gearr{} (\S\ref{sec:gearr}) learns a missing-mass-weighted low-rank residual $\Delta(x)$ for the part that GEAR cannot express, using distillation on a small amount of data.

\subsection{GEAR: Gated Expert Activation Reuse}
\label{sec:gear}

\paragraph{Scaled reuse.}
With its parameters removed, a pruned expert can only be estimated from what remains after pruning. Experts in an MoE LLM form groups with similar activation patterns (\S\ref{sec:introduction}): within the Top-$k$ combination of a token, the output of a pruned expert often points in a direction similar to that of an activated expert and differs mainly in scale~\citep{exfold}. GEAR therefore estimates it by scaling the output of an activated expert:
{\postdisplaypenalty=10000
\begin{equation}
  E_i(x)\approx S_{i\leftarrow j}\,E_j(x),\qquad j\in\mathcal{R}(x).
  \label{eq:scale}
\end{equation}}
We call $S_{i\leftarrow j}$ the \emph{directed scale}, since approximating $i$ with $j$ and $j$ with $i$ generally yield different coefficients. Let $\mathcal{C}_{ij}$ denote the calibration tokens on which experts $i$ and $j$ are both in the Top-$k$ set. One forward pass of the original model over a small calibration corpus yields $S$ and the relative reconstruction error by weighted least squares:
\begin{equation}
\begin{aligned}
  L_{ij}(s)&=\sum_{t\in\mathcal{C}_{ij}}w_t\big\lVert E_i(x_t)-s\,E_j(x_t)\big\rVert_2^2,\\
  S_{i\leftarrow j}&=\arg\min_{s}L_{ij}(s),\quad
  \varepsilon_{i\leftarrow j}=\frac{L_{ij}(S_{i\leftarrow j})}{L_{ij}(0)}.
\end{aligned}
\label{eq:calib}
\end{equation}
Here $\varepsilon$ is the error relative to dropping the term, with $0$ meaning exact reconstruction; both quantities have closed forms.

\paragraph{Best proxy.}
For each pruned expert $i\in\mathcal{D}(x)$, the \emph{best proxy} is the activated expert with the lowest calibrated error among those co-activated with $i$ during calibration:
\begin{equation}
  j_i^*(x)=\arg\min_{j\in\mathcal{R}(x)\cap\mathcal{P}_i}\varepsilon_{i\leftarrow j},
  \label{eq:proxy}
\end{equation}
where $\mathcal{P}_i=\{j:\mathcal{C}_{ij}\neq\emptyset\}$ is the set of experts co-activated with $i$ during calibration. Since $E_{j_i^*}(x)$ is already computed for the current token, reuse adds no expert computation; since the candidates follow the Top-$k$ combination of each token, $\hat{E}_i(x)$ varies from token to token.

\paragraph{Reliability-gated reuse.}
The best proxy is the lowest-error expert among the available candidates, yet its error varies widely across models and expert pairs. A large error injects bias into the residual stream that accumulates over subsequent layers. A fallback that needs no expert computation is the mean output $\mu_i$, the constant that minimizes the expected squared error over the tokens routed to expert $i$, recorded in the same calibration pass. GEAR adopts the best proxy only when its error falls below a threshold $\tau$ and falls back to $\mu_i$ otherwise:
\begin{equation}
  \hat{E}_i(x)=
  \begin{cases}
    S_{i\leftarrow j_i^*}\,E_{j_i^*}(x), & \varepsilon_{i\leftarrow j_i^*}<\tau,\\
    \mu_i, & \text{otherwise}.
  \end{cases}
  \label{eq:gate}
\end{equation}
With $\mathcal{D}_\tau(x)\subseteq\mathcal{D}(x)$ the pruned experts that pass the gate and $y_{\mu}(x)$ the output of Eq.~\eqref{eq:framework} with $\hat{E}_i=\mu_i$ and $\Delta=0$,
\begin{equation}
  \begin{aligned}
  &y_{\mathrm{GEAR}}(x)=y_{\mu}(x)\\
  &\quad+\sum_{i\in\mathcal{D}_\tau(x)}g_i(x)\big[S_{i\leftarrow j_i^*}E_{j_i^*}(x)-\mu_i\big].
  \end{aligned}
  \label{eq:gear}
\end{equation}
GEAR thus builds on the \emph{mean estimate} $y_\mu$, which coincides with MoNE~\citep{mone} on the same mask, and replaces it with the reused output only where the proxy is reliable; $\tau\to0$ recovers the mean estimate, and $\tau=1$ disables the gate.

\paragraph{Threshold calibration.}
The threshold $\tau$ depends on the model and the pruning ratio. With the mask and calibration statistics fixed, we select $\tau^*=\arg\min_{\tau\in\mathcal{G}}\mathrm{NLL}_{\mathrm{dev}}(f_\tau)$ from a small candidate set $\mathcal{G}$, where $f_\tau$ denotes GEAR with threshold $\tau$ and $\mathrm{NLL}_{\mathrm{dev}}$ is the next-token NLL on a development set. The search involves only forward passes. Overall, GEAR requires one calibration pass and a few development passes, and executes no additional experts at inference.

\subsection{\texorpdfstring{GEAR\textsuperscript{2}}{GEAR2}: Missing-Mass Residual}
\label{sec:gearr}

\paragraph{Residual target.}
We write the output of each pruned expert as the GEAR estimate plus a remainder, $E_i(x)=\hat{E}_i(x)+\delta_i(x)$.
When GEAR reuses a proxy, $\hat{E}_i$ is the \emph{shared component} aligned with the proxy and $\delta_i$ is the \emph{non-collinear residual} that no scaling of the proxy can express; when GEAR falls back to the mean, $\delta_i=E_i(x)-\mu_i$ is the input-dependent part that the mean misses. The layer output therefore still lacks $\sum_{i\in\mathcal{D}(x)}g_i(x)\,\delta_i(x)$, which grows with the pruning ratio as each token loses more experts and keeps fewer proxies. This part has no closed form from calibration, so \gearr{} learns only it and leaves the reusable part to GEAR.

\paragraph{Missing-mass weighting.}
We factor the missing part by the \emph{missing mass} $m(x)$:
\begin{equation}
\begin{aligned}
  \sum_{i\in\mathcal{D}(x)}g_i(x)\,\delta_i(x)&=m(x)\sum_{i\in\mathcal{D}(x)}\bar{g}_i(x)\,\delta_i(x),\\
  m(x)=\sum_{i\in\mathcal{D}(x)}g_i(x)&,\quad \bar{g}_i(x)=g_i(x)/m(x).
\end{aligned}
\label{eq:mass}
\end{equation}
The missing mass is the total routing weight that the original router assigns to the pruned experts; it measures how much routed contribution the token loses and is available at inference. The second factor, a routing-weighted average of the residuals, is a function of $x$, which we approximate with a low-rank linear map shared within each layer:
\begin{equation}
  \Delta(x)=m(x)\,BAx,\quad A\in\mathbb{R}^{r\times d},\ B\in\mathbb{R}^{d\times r}.
  \label{eq:residual}
\end{equation}
The correction vanishes when $m(x)=0$ and scales with $m(x)$ otherwise. Each layer adds a single pair of $A$ and $B$ with $2dr$ parameters, independent of the number of pruned experts.

\paragraph{Distillation.}
Rather than fitting $\delta_i$ layer by layer, we train the residuals of all layers end to end against the output distribution of the original model~\citep{kd}. We freeze all original parameters and calibration statistics, initialize $B$ to zero so that training starts from GEAR, and minimize the token-level forward KL divergence:
\begin{equation}
  \mathcal{L}_{\mathrm{KD}}=\frac{1}{\lvert\mathcal{V}\rvert}\sum_{t\in\mathcal{V}}
  \mathrm{KL}\big(p^{(t)}_{\mathrm{orig}}\,\Vert\,p^{(t)}_{\gearr}\big),
  \label{eq:kd}
\end{equation}
where $\mathcal{V}$ is the set of valid tokens. Section~\ref{sec:ablation} validates both choices: the same residual performs better on top of GEAR than on top of the mean estimate, and removing $m(x)$ markedly degrades performance.

\paragraph{Implementation details.}
In Eq.~\eqref{eq:calib}, we set $w_t=\lVert E_i(x_t)\rVert_2$ so that tokens with larger output norms receive larger weights, add $10^{-3}$ to the denominator of $S$ for numerical stability, and clip $S$ to $[-4,4]$. A pruned expert with no candidate in $\mathcal{R}(x)\cap\mathcal{P}_i$ falls back to $\mu_i$, and ties are broken by the original Top-$k$ order. The candidate set of $\tau$ and its rationale are given in Appendix~\ref{sec:appendix-repro}.

\section{Experiments}
\label{sec:experiments}
\input{tables/T1_main_avg.tex}
\input{tables/T2_per_task_50.tex}

\subsection{Experimental Setup}
\label{sec:setup}

\paragraph{Models.}
We evaluate five open-source MoE models from 16B to 106B parameters: Moonlight-16B-A3B~\citep{moonlight}, Qwen3-30B-A3B-Instruct-2507~\citep{qwen3}, DeepSeek-V2-Lite~\citep{deepseekv2}, Gemma-4-26B-A4B~\citep{gemma4}, and GLM-4.5-Air~\citep{glm45}. Their expert configurations are listed in Appendix~\ref{sec:appendix}.

\paragraph{Pruning and baselines.}
We prune 25\% or 50\% of the routed experts in each layer. To examine whether compensation depends on the pruning criterion, we use the REAP mask~\citep{reap} on GLM-4.5-Air and the MoNE mask~\citep{mone} on the other models; Appendix~\ref{sec:appendix-reap} reports REAP masks on three more models, where the gains persist. We compare with methods from three families. Native REAP removes the pruned experts and reroutes tokens among the retained ones. MoNE replaces each pruned expert with its mean output on the same mask and is the direct training-free counterpart of GEAR, while MoRA~\citep{mora} approximates pruned experts with affine transforms of retained ones. HC-SMoE~\citep{hcsmoe} clusters experts and merges each cluster into one. Results of MoRA and HC-SMoE follow \citet{mora} and are shown in italics.

\paragraph{Implementation details.}
Following common practice~\citep{sparsegpt,wanda}, we calibrate on C4~\citep{c4}: 100 packed sequences of 4{,}096 tokens from the training set for the statistics, and 16 documents for selecting $\tau$. The residual of \gearr{} has rank 32 and is distilled for 64 steps on 64 sequences of 512 tokens from C4. Further details are given in Appendix~\ref{sec:appendix}.

\paragraph{Evaluation.}
We use lm-evaluation-harness~\citep{lmeval} to evaluate nine zero-shot tasks: ARC-Challenge and ARC-Easy~\citep{arc}, BoolQ~\citep{boolq}, HellaSwag~\citep{hellaswag}, MMLU~\citep{mmlu}, OpenBookQA~\citep{openbookqa}, PIQA~\citep{piqa}, RTE~\citep{rte,superglue}, and WinoGrande~\citep{winogrande}. We report acc\_norm for ARC, HellaSwag, OpenBookQA, and PIQA and acc for the others, and we report the unweighted average over the nine tasks. Bold marks the best pruned result in each column of a model.

\subsection{Main Results}
\label{sec:main-results}

Table~\ref{tab:main} compares the nine-task average accuracy on five models at two pruning ratios. Our methods achieve the best pruned result in all 10 settings, 9 of them by \gearr{}, and \gearr{} outperforms both same-mask MoNE and native REAP in every setting. At 50\% pruning, \gearr{} improves over MoNE by 0.96--3.44 points. Without any training, GEAR already surpasses MoNE in 9 of the 10 settings, by 1.49 and 1.16 points on DeepSeek-V2-Lite and Qwen3 at 50\%. \gearr{} also exceeds the substitution method MoRA and the merging method HC-SMoE on every model they report, by 2.75--6.41 and 3.31--25.34 points at 50\% pruning.

\begin{figure*}[t]
  \centering
  \includegraphics[width=0.9\textwidth]{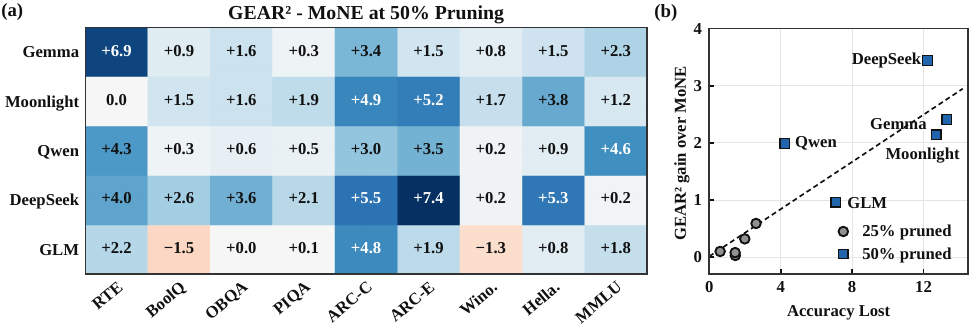}
  \caption{(a)~Per-task gain of \gearr{} over MoNE at 50\% pruning (pp). (b)~Accuracy lost to pruning versus the gain of \gearr{} over MoNE: the gain grows with the loss. Dashed line: least-squares fit.}
  \label{fig:task-gain}
\end{figure*}

The gain from compensation grows with the damage caused by pruning. Across the 10 settings, the accuracy lost to pruning and the gain of \gearr{} over MoNE reach a Spearman correlation of 0.90 (Figure~\ref{fig:task-gain}b): at 25\%, pruning costs only 0.6--2.6 points, leaving little to recover, whereas at 50\% the loss rises to 4.2--13.3 points and the gain widens accordingly. Per task at 50\% (Figure~\ref{fig:task-gain}a and Table~\ref{tab:pertask}), the largest gains appear on ARC-Challenge and ARC-Easy, which pruning damages by 16.5 and 16.9 points on average: \gearr{} recovers 4.3 and 3.9 points and improves on all five models. Overall, \gearr{} improves over MoNE on 41 of the 45 model--task pairs, and on every model for five of the nine tasks. By source, the residual improves over GEAR on 41 pairs and reuse improves over MoNE on 30: reuse recovers what the activated experts can express, and the residual recovers the rest. These gains are significant under a paired bootstrap and robust to seeds and pruning masks (Appendix~\ref{sec:appendix-ci}--\ref{sec:appendix-robust}).

\input{tables/T2_gate_ablation.tex}
\input{tables/T3_residual_ablation.tex}

\subsection{Ablation Study}
\label{sec:ablation}

This section examines the key design of each method: the reliability gate of GEAR, and the base and weighting of the residual in \gearr{}.

\paragraph{Reliability gate.}
The gate must identify reliable proxies rather than merely reduce reuse: masking the pairs $(i\leftarrow j)$ whose error exceeds $\tau$ does both at once. To separate the two, on Gemma-4 at 50\% we keep the number of masked pairs equal to that of GEAR and change only which pairs are masked, either at random or in reverse order of $\varepsilon$, which masks the lowest-error pairs and keeps the highest. Table~\ref{tab:gate} shows that two random masks and the reverse mask fall 0.96, 1.57, and 1.00 points below GEAR, all below the mean estimate. The same reuse frequency thus yields opposite outcomes depending on which pairs are kept, so the gain of GEAR comes from keeping reliable proxies according to $\varepsilon$ rather than from reusing less. Without the gate, always reusing the best proxy falls below the mean estimate by 0.63 points on Gemma-4 and by 5.01 points on Moonlight at 50\%, while GEAR with the gate exceeds it on both (Table~\ref{tab:main}). Unreliable reuse is more harmful than no reuse: $\mu_i$ is the minimum-error constant and introduces no input-dependent bias, whereas a biased proxy output enters the residual stream at every layer and accumulates. The gate is also easy to set: every $\tau<1$ in $\mathcal{G}$ recovers 4.86--5.30 points over ungated reuse on Moonlight (Appendix~\ref{sec:appendix-robust}), and replacing the $\varepsilon$-selected proxy with an oracle choice reduces the reconstruction error by at most 0.17\% (Appendix~\ref{sec:appendix-proxy}).

\paragraph{Residual design.}
The residual of \gearr{} is built on the estimate of GEAR and weighted by the missing mass $m(x)$ (Eq.~\eqref{eq:mass}). Table~\ref{tab:residual} tests both on Qwen3 at 50\% under the same parameter count and training budget. For the base, the same residual built on zero or on the mean estimate scores 0.62 and 0.63 points lower than on GEAR, although it then faces a larger remaining error. This is consistent with the decomposition in \S\ref{sec:gearr}: a low-rank map is better suited to the non-collinear remainder after reuse than to the full contribution, so reuse and the residual are complementary. For the weighting, replacing $m(x)$ with the constant 1 lowers the average to 68.62, below the mean estimate without any residual. With $m(x)$, the correction vanishes on tokens that lose no routing weight and grows with the weight a token loses; a constant weight applies the same correction to every token, including those whose Top-$k$ set contains no pruned expert.

\begin{figure}[t]
  \centering
  \includegraphics[width=\columnwidth]{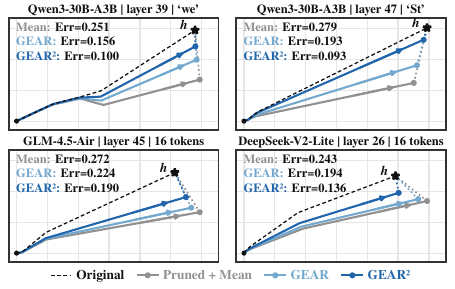}
  \caption{Layer-wise hidden states at 50\% pruning, projected onto two dimensions. Qualitative examples, selected by the error reduction of \gearr{}. Err: normalized $L_2$ error of the hidden state relative to the original model at the shown layer.}
  \label{fig:traj}
\end{figure}

\subsection{Representation Analysis}
\label{sec:representation}

To illustrate how compensation acts on the forward computation, Figure~\ref{fig:traj} projects the layer-wise hidden states of the original and compensated models onto two dimensions, with examples from Qwen3, GLM-4.5-Air, and DeepSeek-V2-Lite at 50\% pruning. After pruning, the trajectory with mean compensation drifts away from the original model layer by layer. GEAR recovers the pruned contributions from the experts already executed for the current token and pulls the trajectory back toward the original. The residual of \gearr{} further corrects what reuse cannot express and brings the endpoint closer to the original model.

\input{tables/T4_budget.tex}

\subsection{Overhead}
\label{sec:overhead}

Table~\ref{tab:budget} summarizes the cost on Qwen3 at 50\% pruning: GEAR needs no training, and \gearr{} trains only 64 steps and adds about 0.08\,GiB of storage, under 0.3\% of the pruned model. At inference, both only reuse expert outputs that have already been computed; the extra table lookups, scalar scaling, weighted summation, and low-rank projection can be fused into the output aggregation kernel of the MoE layer. In our unfused implementation (batch size 1, 128 prompt and 32 generated tokens), MoNE, GEAR, and \gearr{} take 2.54, 3.15, and 3.47 seconds, respectively.

\section{Conclusion}
\label{sec:conclusion}

This paper shows that the contribution of pruned experts can be recovered from the outputs a pruned MoE already computes. GEAR reuses the scaled outputs of experts already activated for the same token behind a reliability gate, without training or additional expert computation, and \gearr{} adds a missing-mass-weighted low-rank residual distilled for only 64 steps. Both leave the router and retained experts untouched and apply on top of any pruning mask. On five MoE models, \gearr{} outperforms MoNE and REAP in every setting, and the more aggressive the pruning, the larger the gain. Recovery after pruning need not rely on retraining: by making full use of what a pruned model already computes, the accuracy loss from pruning can be narrowed at very low cost.

\section*{Limitations}

Our evaluation covers nine zero-shot understanding and reasoning tasks on MoE models from 16B to 106B parameters. Generative and long-context tasks and larger models are left for future work, where calibration and distillation data matched to the target domain are a natural extension. Our reference implementation is not yet fused into the output aggregation of the MoE layer, which would further reduce the inference overhead.

\section*{Ethical Considerations}

This work reduces the memory and computation needed to deploy MoE models at little training cost. All models and datasets we use are publicly available and used under their licenses.


\ifhasreferences
\bibliography{refs}
\fi

\clearpage
\appendix
\raggedbottom

\section{Implementation Details}
\label{sec:appendix}
\label{sec:appendix-repro}

\paragraph{Models.} Moonlight-16B-A3B and DeepSeek-V2-Lite activate 6 of 64 routed experts alongside 2 shared experts in each MoE layer; Qwen3 and Gemma-4 activate 8 of 128 routed experts; GLM-4.5-Air activates 8 of 128 routed experts alongside 1 shared expert.

\paragraph{Calibration.} All statistics of a layer are collected in a single forward pass of the original model over 100 packed sequences of 4{,}096 tokens from the C4 training set. The mean output $\mu_i$ is averaged over the tokens routed to expert $i$, and $S_{i\leftarrow j}$ and $\varepsilon_{i\leftarrow j}$ are computed over the tokens $\mathcal{C}_{ij}$ on which $i$ and $j$ are co-activated.

\paragraph{Threshold selection.} We search $\tau$ over $\mathcal{G}=\{0.90, 0.95, 0.98, 0.99, 1\}$, where $\tau=1$ disables the gate. Since $\varepsilon$ measures the error relative to dropping the term, a proxy with $\varepsilon$ close to $1$ offers little over the mean estimate, so the candidates concentrate near $1$. For each model and pruning ratio, we select the threshold with the lowest next-token NLL on 16 development documents from C4 and break ties toward the larger $\tau$, which keeps more reuse. The search involves only forward passes. The selection transfers to held-out text: on Gemma-4 at 50\% pruning, the threshold selected on the development documents ($\tau=0.98$) also yields the lowest NLL on 16 further C4 documents.

\paragraph{Training.} The residual is trained with AdamW~\citep{adamw} at a learning rate of $5\times10^{-4}$, with gradient clipping at 1.0 and a distillation temperature of 1.

\paragraph{Residual size.} With rank 32, the residual of \gearr{} adds 3.4M parameters to Moonlight and DeepSeek-V2-Lite, 6.3M to Qwen3, 5.4M to Gemma-4, and 11.8M to GLM-4.5-Air, about 0.02\% of each model or less.

\paragraph{Evaluation.} All tasks are evaluated zero-shot with lm-evaluation-harness. MMLU accuracy is averaged over all its questions, and the reported average weights the nine tasks equally.

\section{Additional Results}
\label{sec:appendix-results}

\subsection{Per-Task Results at 25\% Pruning}
\label{sec:appendix-pertask25}
Table~\ref{tab:appendix-pertask25} reports the per-task accuracy at 25\% pruning in the same format as Table~\ref{tab:pertask}. \gearr{} stays within 2.1 points of the original model on every model and attains the best pruned average on four of the five models.
\input{tables/A_per_task_25.tex}

\subsection{Statistical Significance}
\label{sec:appendix-ci}
We estimate 95\% confidence intervals of the nine-task average difference with a paired bootstrap over test questions, resampling within each task (4{,}000 resamples). At 50\% pruning, the improvements of \gearr{} over MoNE, REAP, and GEAR are significant on all five models (Table~\ref{tab:appendix-ci}). With disjoint question halves for estimating pruning damage and recovery gain, cross-fitted Spearman correlations remain positive across the nine available model--pruning settings (median 0.87; 2.5--97.5\% split quantiles [0.72, 0.95]) and across the 45 model--task pairs at 50\% pruning (median 0.45; [0.34, 0.58]).
\input{tables/A_paired_ci.tex}

\subsection{Results with REAP Masks}
\label{sec:appendix-reap}
To test whether compensation depends on the pruning criterion, we replace the MoNE mask with the REAP mask on Qwen3, DeepSeek-V2-Lite, and Gemma-4 at 50\% pruning and follow the same procedure otherwise. GEAR and \gearr{} improve over the mean estimate by 1.16--2.44 and 1.83--3.61 points, respectively (Table~\ref{tab:appendix-reap}), so the gains do not rely on a specific mask.
\input{tables/A_reap_mask.tex}
Per task, \gearr{} improves over the mean estimate on 23 of the 27 model--task pairs (Table~\ref{tab:appendix-reap-pertask}).
\input{tables/A_reap_mask_pertask.tex}

\subsection{Robustness to Seed, Rank, and $\tau$}
\label{sec:appendix-robust}
Table~\ref{tab:appendix-robust} retrains the residual of \gearr{} with three seeds on Moonlight and DeepSeek-V2-Lite at 50\% pruning: every seed improves over MoNE by at least 1.83 and 2.80 points, respectively, and rank 16 retains most of the gain.
\input{tables/A_robustness.tex}
Table~\ref{tab:appendix-tau} varies the threshold of GEAR on Moonlight: every $\tau<1$ in $\mathcal{G}$ recovers 4.86--5.30 points over ungated reuse ($\tau=1$), so the gate does not hinge on a finely tuned $\tau$.
\input{tables/A_tau_sweep.tex}

\subsection{Robustness across Calibration Batches}
\label{sec:appendix-batches}
On GLM-4.5-Air, we repeat calibration and residual training with three calibration batches. \gearr{} outperforms REAP on all six combinations of batch and pruning ratio, by 0.82--1.49 points at 50\% pruning (Table~\ref{tab:appendix-batches}).
\input{tables/A_glm_batches.tex}

\section{Additional Analysis}
\label{sec:appendix-analysis}

\subsection{Proxy Selection}
\label{sec:appendix-proxy}
We examine the two choices behind the best proxy on Moonlight and DeepSeek-V2-Lite at 50\% pruning, measuring the reconstruction error of the pruned contribution on 8 sequences of 512 tokens. First, enlarging the candidate pool from the co-activated experts to all retained experts changes this error by about 0.1\%, while it would require 18.0\% and 30.2\% more expert evaluations, respectively. Second, replacing the $\varepsilon$-selected proxy with an oracle that picks, for every slot, the candidate with the smallest true error reduces the per-slot error by only 0.17\% and 0.04\%. Restricting reuse to co-activated experts and selecting proxies by $\varepsilon$ is therefore nearly optimal at no extra cost.

\subsection{Hidden-State Trajectories}
\label{sec:appendix-traj}
The examples in Figure~\ref{fig:traj} (\S\ref{sec:representation}) are chosen for illustration; Table~\ref{tab:appendix-traj} confirms the trend over all token windows. Over 150 windows per model, \gearr{} lowers the average hidden-state error of the mean estimate on every model and is closer to the original model in 111, 142, and 99 windows on Moonlight, DeepSeek-V2-Lite, and GLM-4.5-Air, respectively.
\input{tables/A_traj_windows.tex}

\end{document}

%% file: tables/table_style.tex
\newcolumntype{Y}{>{\centering\arraybackslash}X}
\definecolor{GearBlue}{HTML}{2166AC}
\definecolor{GearGray}{HTML}{919191}
\definecolor{GearHeaderGray}{HTML}{FFFFFF}
\definecolor{GearGroupGray}{HTML}{F2F2F2}
\definecolor{GearMeanFill}{HTML}{FFFFFF}
\definecolor{GearReuseFill}{HTML}{FFFFFF}
\definecolor{GearFinalFill}{HTML}{DBE9F3}
\newcommand{\gearBest}[2][white]{#2}
\newcommand{\gearTableName}{{\normalfont\itshape GEAR}}
\newcommand{\gearTableNameSquared}{{\normalfont\itshape GEAR\textsuperscript{2}}}
\newcommand{\gearMethod}[1]{{\leavevmode\color{GearBlue}\bfseries\boldmath #1}}
\newcommand{\gearHeaderCell}[1]{{\bfseries\boldmath #1}}
\newcommand{\gearTableSetup}{%
  \arrayrulecolor{black}%
  \setlength{\heavyrulewidth}{0.7pt}%
  \setlength{\lightrulewidth}{0.4pt}%
  \setlength{\cmidrulewidth}{0.35pt}%
  \setlength{\aboverulesep}{1.8pt}%
  \setlength{\belowrulesep}{1.8pt}%
  \renewcommand{\arraystretch}{0.9}%
}

%% file: tables/T1_main_avg.tex
\begin{table*}[t]
\centering\small
\setlength{\tabcolsep}{4.2pt}
\gearTableSetup
\begin{tabularx}{\linewidth}{lYYYYYYYYYY}
\toprule
& \multicolumn{2}{c}{\gearHeaderCell{Moonlight}} & \multicolumn{2}{c}{\gearHeaderCell{Qwen3}} & \multicolumn{2}{c}{\gearHeaderCell{DeepSeek}} & \multicolumn{2}{c}{\gearHeaderCell{Gemma-4}} & \multicolumn{2}{c}{\gearHeaderCell{GLM}} \\
\cmidrule(lr){2-3} \cmidrule(lr){4-5} \cmidrule(lr){6-7} \cmidrule(lr){8-9} \cmidrule(lr){10-11}
\gearHeaderCell{Method} & 25\% & 50\% & 25\% & 50\% & 25\% & 50\% & 25\% & 50\% & 25\% & 50\% \\
\midrule
Original & \multicolumn{2}{c}{69.99} & \multicolumn{2}{c}{74.48} & \multicolumn{2}{c}{65.93} & \multicolumn{2}{c}{75.10} & \multicolumn{2}{c}{75.76} \\
HC-SMoE & \textit{56.87} & \textit{44.54} & \textit{64.98} & \textit{46.90} & \textit{62.66} & \textit{53.84} & -- & -- & -- & -- \\
MoRA & \textit{67.48} & \textit{55.36} & \textit{72.50} & \textit{65.83} & \textit{63.68} & \textit{54.40} & -- & -- & -- & -- \\
\midrule
REAP & 66.68 & 55.37 & 73.56 & 67.41 & 63.19 & 54.36 & 69.61 & 59.20 & 74.33 & 68.81 \\
MoNE & 67.37 & 56.68 & 73.87 & 70.25 & 64.47 & 53.71 & 73.11 & 62.39 & 74.31 & 68.67 \\
\gearTableName{} & 67.66 & 56.98 & 73.91 & 71.41 & 64.49 & 55.20 & 73.25 & 62.98 & \textbf{74.43} & 68.29 \\
\rowcolor{GearFinalFill}\gearMethod{\gearTableNameSquared{}} & \textbf{67.96} & \textbf{59.09} & \textbf{73.97} & \textbf{72.24} & \textbf{64.50} & \textbf{57.15} & \textbf{73.43} & \textbf{64.53} & 74.39 & \textbf{69.63} \\
\bottomrule
\end{tabularx}
\caption{Nine-task average accuracy (\%) after pruning 25\% or 50\% of routed experts.}
\label{tab:main}
\end{table*}

%% file: tables/T2_per_task_50.tex
\begin{table*}[t]
\centering\small
\setlength{\tabcolsep}{5pt}
\gearTableSetup
\renewcommand{\arraystretch}{0.9}
\begin{tabularx}{\linewidth}{@{\hspace{2pt}}c@{\hspace{4pt}}lYYYYYYYYYY}
\toprule
& \gearHeaderCell{Method} & \gearHeaderCell{RTE} & \gearHeaderCell{BoolQ} & \gearHeaderCell{OBQA} & \gearHeaderCell{PIQA} & \gearHeaderCell{ARC-C} & \gearHeaderCell{ARC-E} & \gearHeaderCell{Wino.} & \gearHeaderCell{Hella.} & \gearHeaderCell{MMLU} & \gearHeaderCell{Avg.} \\
\midrule
 & Original & 66.06 & 80.24 & 45.20 & 80.79 & 57.94 & 82.41 & 71.43 & 78.36 & 67.48 & 69.99 \\
 & HC-SMoE & \textit{53.43} & \textit{61.31} & \textit{27.80} & \textit{62.73} & \textit{26.02} & \textit{50.63} & \textit{54.06} & \textit{34.96} & \gearBest[white]{\textit{\textbf{29.93}}} & \textit{44.54} \\
 & MoRA & \gearBest[white]{\textit{\textbf{61.73}}} & \textit{70.92} & \textit{35.40} & \textit{77.31} & \textit{36.43} & \textit{65.32} & \textit{61.01} & \textit{61.47} & \textit{28.66} & \textit{55.36} \\
 & REAP & 46.21 & 56.51 & \gearBest[white]{\textbf{40.80}} & 77.42 & \gearBest[white]{\textbf{42.83}} & \gearBest[white]{\textbf{66.25}} & 70.80 & 72.76 & 24.76 & 55.37 \\
 & MoNE & 59.57 & 72.26 & 39.20 & 78.29 & 36.43 & 59.89 & 69.53 & 72.01 & 22.97 & 56.68 \\
 & \gearTableName{} & 60.65 & 72.11 & 39.60 & 78.51 & 36.26 & 60.06 & 70.40 & 72.22 & 22.97 & 56.98 \\
\rowcolor{GearFinalFill}\cellcolor{white}\multirow{-7}{*}{\rotatebox[origin=c]{90}{\textit{Moonlight}}} & \gearMethod{\gearTableNameSquared{}} & 59.57 & \gearBest[GearFinalFill]{\textbf{73.79}} & \gearBest[GearFinalFill]{\textbf{40.80}} & \gearBest[GearFinalFill]{\textbf{80.14}} & 41.30 & 65.07 & \gearBest[GearFinalFill]{\textbf{71.19}} & \gearBest[GearFinalFill]{\textbf{75.76}} & 24.16 & \gearBest[GearFinalFill]{\textbf{59.09}} \\
\midrule
 & Original & 76.90 & 88.56 & 44.80 & 80.09 & 62.63 & 83.84 & 73.64 & 79.76 & 80.11 & 74.48 \\
 & HC-SMoE & \textit{58.12} & \textit{72.29} & \textit{27.40} & \textit{56.47} & \textit{30.03} & \textit{47.39} & \textit{53.43} & \textit{35.61} & \textit{41.33} & \textit{46.90} \\
 & MoRA & \textit{75.09} & \textit{82.51} & \textit{41.80} & \textit{80.47} & \textit{46.25} & \textit{69.07} & \textit{68.98} & \textit{75.20} & \textit{53.14} & \textit{65.83} \\
 & REAP & 81.23 & 84.89 & 38.40 & 77.15 & 52.30 & 71.17 & 69.85 & 70.05 & 61.64 & 67.41 \\
 & MoNE & 79.78 & 87.89 & 43.80 & 81.18 & 53.50 & 74.92 & 74.03 & 78.27 & 58.87 & 70.25 \\
 & \gearTableName{} & 79.78 & \gearBest[white]{\textbf{88.35}} & 44.20 & 81.45 & 56.14 & 77.02 & 74.11 & 78.49 & 63.10 & 71.41 \\
\rowcolor{GearFinalFill}\cellcolor{white}\multirow{-7}{*}{\rotatebox[origin=c]{90}{\textit{Qwen3}}} & \gearMethod{\gearTableNameSquared{}} & \gearBest[GearFinalFill]{\textbf{84.12}} & 88.20 & \gearBest[GearFinalFill]{\textbf{44.40}} & \gearBest[GearFinalFill]{\textbf{81.66}} & \gearBest[GearFinalFill]{\textbf{56.48}} & \gearBest[GearFinalFill]{\textbf{78.45}} & \gearBest[GearFinalFill]{\textbf{74.27}} & \gearBest[GearFinalFill]{\textbf{79.13}} & \gearBest[GearFinalFill]{\textbf{63.48}} & \gearBest[GearFinalFill]{\textbf{72.24}} \\
\midrule
 & Original & 61.01 & 79.72 & 43.80 & 80.03 & 48.55 & 76.22 & 71.03 & 77.86 & 55.11 & 65.93 \\
 & HC-SMoE & \textit{57.04} & \textit{66.51} & \textit{34.56} & \textit{71.58} & \textit{33.69} & \textit{60.57} & \textit{65.56} & \textit{60.75} & \gearBest[white]{\textit{\textbf{34.32}}} & \textit{53.84} \\
 & MoRA & \textit{51.63} & \textit{58.96} & \gearBest[white]{\textit{\textbf{42.20}}} & \textit{77.80} & \textit{38.65} & \textit{63.17} & \textit{62.51} & \textit{71.47} & \textit{23.25} & \textit{54.40} \\
 & REAP & 54.51 & 59.17 & 40.00 & 76.28 & 40.78 & 61.83 & 63.69 & 69.31 & 23.63 & 54.36 \\
 & MoNE & 54.15 & 66.21 & 36.80 & 76.28 & 35.58 & 56.06 & 67.48 & 67.84 & 22.96 & 53.71 \\
 & \gearTableName{} & 55.96 & 65.47 & 38.60 & 77.37 & 38.57 & 60.48 & 66.69 & 70.67 & 23.02 & 55.20 \\
\rowcolor{GearFinalFill}\cellcolor{white}\multirow{-7}{*}{\rotatebox[origin=c]{90}{\textit{DeepSeek}}} & \gearMethod{\gearTableNameSquared{}} & \gearBest[GearFinalFill]{\textbf{58.12}} & \gearBest[GearFinalFill]{\textbf{68.78}} & 40.40 & \gearBest[GearFinalFill]{\textbf{78.35}} & \gearBest[GearFinalFill]{\textbf{41.13}} & \gearBest[GearFinalFill]{\textbf{63.51}} & \gearBest[GearFinalFill]{\textbf{67.72}} & \gearBest[GearFinalFill]{\textbf{73.17}} & 23.19 & \gearBest[GearFinalFill]{\textbf{57.15}} \\
\midrule
 & Original & 67.15 & 85.08 & 48.20 & 83.73 & 68.69 & 89.56 & 76.01 & 83.20 & 74.31 & 75.10 \\
 & REAP & 58.84 & 77.74 & 41.20 & 79.54 & 42.15 & 64.52 & 66.14 & 71.71 & \gearBest[white]{\textbf{30.97}} & 59.20 \\
 & MoNE & 60.65 & 81.13 & 47.00 & 82.10 & 42.41 & 67.26 & 74.51 & 80.00 & 26.50 & 62.39 \\
 & \gearTableName{} & 62.82 & 79.14 & 47.60 & 82.05 & 44.80 & 68.56 & 74.35 & 80.48 & 27.00 & 62.98 \\
\rowcolor{GearFinalFill}\cellcolor{white}\multirow{-5}{*}{\rotatebox[origin=c]{90}{\textit{Gemma-4}}} & \gearMethod{\gearTableNameSquared{}} & \gearBest[GearFinalFill]{\textbf{67.51}} & \gearBest[GearFinalFill]{\textbf{82.05}} & \gearBest[GearFinalFill]{\textbf{48.60}} & \gearBest[GearFinalFill]{\textbf{82.43}} & \gearBest[GearFinalFill]{\textbf{45.82}} & \gearBest[GearFinalFill]{\textbf{68.73}} & \gearBest[GearFinalFill]{\textbf{75.30}} & \gearBest[GearFinalFill]{\textbf{81.54}} & 28.82 & \gearBest[GearFinalFill]{\textbf{64.53}} \\
\midrule
 & Original & 75.09 & 88.23 & 48.20 & 83.46 & 61.86 & 82.66 & 77.51 & 85.87 & 78.93 & 75.76 \\
 & REAP & 62.09 & 84.28 & \gearBest[white]{\textbf{45.40}} & 80.90 & 52.22 & 73.74 & 77.19 & \gearBest[white]{\textbf{82.16}} & 61.29 & 68.81 \\
 & MoNE & 66.43 & \gearBest[white]{\textbf{87.19}} & 43.40 & 81.12 & 49.23 & 72.05 & \gearBest[white]{\textbf{78.14}} & 80.85 & 59.66 & 68.67 \\
 & \gearTableName{} & 63.18 & 86.18 & 42.60 & 81.07 & 50.85 & 72.81 & 76.87 & 80.82 & 60.20 & 68.29 \\
\rowcolor{GearFinalFill}\cellcolor{white}\multirow{-5}{*}{\rotatebox[origin=c]{90}{\textit{GLM-4.5}}} & \gearMethod{\gearTableNameSquared{}} & \gearBest[GearFinalFill]{\textbf{68.59}} & 85.66 & 43.40 & \gearBest[GearFinalFill]{\textbf{81.18}} & \gearBest[GearFinalFill]{\textbf{54.01}} & \gearBest[GearFinalFill]{\textbf{73.99}} & 76.80 & 81.63 & \gearBest[GearFinalFill]{\textbf{61.44}} & \gearBest[GearFinalFill]{\textbf{69.63}} \\
\bottomrule
\end{tabularx}
\caption{Accuracy (\%) on nine zero-shot tasks at 50\% pruning.}
\label{tab:pertask}
\end{table*}

%% file: tables/T2_gate_ablation.tex
\begin{table}[t]
\centering\footnotesize
\setlength{\tabcolsep}{3.4pt}
\gearTableSetup
\begin{tabularx}{\linewidth}{lYYY}
\toprule
\gearHeaderCell{Method} & \gearHeaderCell{Mask} & \gearHeaderCell{Avg.} & \gearHeaderCell{$\Delta$} \\
\midrule
Mean & -- & 62.39 & $-0.59$ \\
No gate & No & 61.76 & $-1.22$ \\
Random \#1 & Yes & 62.02 & $-0.96$ \\
Random \#2 & Yes & 61.41 & $-1.57$ \\
Reverse $\varepsilon$ & Yes & 61.98 & $-1.00$ \\
\rowcolor{GearFinalFill}\gearMethod{\gearTableName{}} & Yes & \textbf{62.98} & ref. \\
\bottomrule
\end{tabularx}
\caption{Reliability-gate ablation on Gemma-4 at 50\% pruning. Mean: every pruned expert falls back to $\mu_i$; masked variants mask as many pairs as GEAR. $\Delta$: relative to GEAR.}
\label{tab:gate}
\end{table}

%% file: tables/T3_residual_ablation.tex
\begin{table}[t]
\centering\footnotesize
\setlength{\tabcolsep}{3.4pt}
\gearTableSetup
\begin{tabularx}{\linewidth}{lYYYY}
\toprule
\gearHeaderCell{Base} & \gearHeaderCell{Res.} & \gearHeaderCell{$m(x)$} & \gearHeaderCell{Avg.} & \gearHeaderCell{$\Delta$} \\
\midrule
Mean & -- & -- & 70.25 & $-1.99$ \\
\gearTableName{} & -- & -- & 71.41 & $-0.83$ \\
\midrule
Zero & \ding{51} & \ding{51} & 71.62 & $-0.62$ \\
Mean & \ding{51} & \ding{51} & 71.61 & $-0.63$ \\
\gearTableName{} & \ding{51} & \ding{55} & 68.62 & $-3.62$ \\
\rowcolor{GearFinalFill}\gearMethod{\gearTableName{}} & \ding{51} & \ding{51} & \textbf{72.24} & ref. \\
\bottomrule
\end{tabularx}
\caption{Residual ablation on Qwen3 at 50\% pruning. Res.: low-rank residual; \ding{55}: $m(x)$ replaced by 1. Last row: \gearr{}. $\Delta$: relative to \gearr{}.}
\label{tab:residual}
\end{table}

%% file: tables/T4_budget.tex
\begin{table}[t]
\centering\footnotesize
\setlength{\tabcolsep}{3.0pt}
\gearTableSetup
\begin{tabularx}{\linewidth}{lccY}
\toprule
\multicolumn{1}{l}{\gearHeaderCell{\strut Metric}} & \multicolumn{1}{c}{\gearHeaderCell{\strut MoNE}} & \multicolumn{1}{c}{\gearHeaderCell{\strut\gearTableName{}}} & \multicolumn{1}{c}{\gearHeaderCell{\strut\gearMethod{\gearTableNameSquared{}}}} \\
\midrule
Training & -- & -- & \mbox{64 steps, 32K tokens} \\
Stats. & $\mu_i$ & $\mu_i,\,S,\,\varepsilon$ & $\mu_i,\,S,\,\varepsilon$ \\
Storage (GiB) & 0.047 & 0.056 & 0.079 \\
Extra experts & 0 & 0 & 0 \\
Latency (s) & 2.54 & 3.15 & 3.47 \\
\bottomrule
\end{tabularx}
\caption{Recovery cost and inference overhead on Qwen3 at 50\% pruning; the pruned model occupies 29.9\,GiB in bf16.}
\label{tab:budget}
\end{table}

%% file: tables/A_per_task_25.tex
\begin{table*}[!t]
\centering\small
\setlength{\tabcolsep}{5pt}
\gearTableSetup
\renewcommand{\arraystretch}{0.9}
\begin{tabularx}{\linewidth}{@{\hspace{2pt}}c@{\hspace{4pt}}lYYYYYYYYYY}
\toprule
& \gearHeaderCell{Method} & \gearHeaderCell{RTE} & \gearHeaderCell{BoolQ} & \gearHeaderCell{OBQA} & \gearHeaderCell{PIQA} & \gearHeaderCell{ARC-C} & \gearHeaderCell{ARC-E} & \gearHeaderCell{Wino.} & \gearHeaderCell{Hella.} & \gearHeaderCell{MMLU} & \gearHeaderCell{Avg.} \\
\midrule
 & Original & 66.06 & 80.24 & 45.20 & 80.79 & 57.94 & 82.41 & 71.43 & 78.36 & 67.48 & 69.99 \\
 & HC-SMoE & \textit{59.57} & \textit{70.34} & \textit{36.40} & \textit{68.99} & \textit{41.72} & \textit{70.45} & \textit{56.59} & \textit{53.93} & \textit{53.80} & \textit{56.87} \\
 & MoRA & \gearBest[white]{\textit{\textbf{70.76}}} & \gearBest[white]{\textit{\textbf{79.57}}} & \textit{43.00} & \textit{80.79} & \textit{53.15} & \textit{79.12} & \textit{69.38} & \textit{74.76} & \gearBest[white]{\textit{\textbf{56.79}}} & \textit{67.48} \\
 & REAP & 55.60 & 79.11 & 45.60 & 81.12 & 54.27 & 78.28 & \gearBest[white]{\textbf{72.61}} & 78.11 & 55.39 & 66.68 \\
 & MoNE & 62.09 & 79.24 & 45.60 & 81.39 & 55.46 & 79.46 & 71.82 & 78.02 & 53.22 & 67.37 \\
 & \gearTableName{} & 63.18 & 79.14 & \gearBest[white]{\textbf{46.00}} & 81.28 & 55.46 & 79.55 & 71.67 & 77.68 & 54.99 & 67.66 \\
\rowcolor{GearFinalFill}\cellcolor{white}\multirow{-7}{*}{\rotatebox[origin=c]{90}{\textit{Moonlight}}} & \gearMethod{\gearTableNameSquared{}} & 62.45 & 79.20 & 45.40 & \gearBest[GearFinalFill]{\textbf{81.45}} & \gearBest[GearFinalFill]{\textbf{56.40}} & \gearBest[GearFinalFill]{\textbf{80.35}} & 71.74 & \gearBest[GearFinalFill]{\textbf{78.35}} & 56.29 & \gearBest[GearFinalFill]{\textbf{67.96}} \\
\midrule
 & Original & 76.90 & 88.56 & 44.80 & 80.09 & 62.63 & 83.84 & 73.64 & 79.76 & 80.11 & 74.48 \\
 & HC-SMoE & \gearBest[white]{\textit{\textbf{80.50}}} & \textit{83.55} & \textit{38.00} & \textit{72.14} & \textit{47.18} & \textit{71.84} & \textit{65.27} & \textit{61.98} & \textit{64.35} & \textit{64.98} \\
 & MoRA & \textit{76.17} & \textit{88.81} & \gearBest[white]{\textit{\textbf{44.80}}} & \gearBest[white]{\textit{\textbf{81.18}}} & \textit{57.42} & \textit{80.18} & \textit{72.38} & \textit{79.28} & \textit{72.32} & \textit{72.50} \\
 & REAP & 74.73 & \gearBest[white]{\textbf{89.08}} & 44.40 & 80.85 & \gearBest[white]{\textbf{63.40}} & 83.54 & 73.72 & 77.82 & 74.54 & 73.56 \\
 & MoNE & 75.81 & 88.23 & 44.40 & 80.90 & 61.69 & 84.39 & 74.35 & 79.72 & 75.35 & 73.87 \\
 & \gearTableName{} & 76.53 & 88.04 & 43.60 & 80.79 & 62.03 & \gearBest[white]{\textbf{84.43}} & 74.03 & 79.70 & \gearBest[white]{\textbf{76.04}} & 73.91 \\
\rowcolor{GearFinalFill}\cellcolor{white}\multirow{-7}{*}{\rotatebox[origin=c]{90}{\textit{Qwen3}}} & \gearMethod{\gearTableNameSquared{}} & 77.26 & 88.10 & 43.60 & 80.85 & 61.77 & 84.01 & \gearBest[GearFinalFill]{\textbf{74.59}} & \gearBest[GearFinalFill]{\textbf{79.73}} & 75.83 & \gearBest[GearFinalFill]{\textbf{73.97}} \\
\midrule
 & Original & 61.01 & 79.72 & 43.80 & 80.03 & 48.55 & 76.22 & 71.03 & 77.86 & 55.11 & 65.93 \\
 & HC-SMoE & \textit{61.37} & \textit{73.37} & \textit{40.60} & \textit{78.78} & \textit{44.88} & \textit{72.10} & \gearBest[white]{\textit{\textbf{71.03}}} & \textit{74.49} & \textit{47.29} & \textit{62.66} \\
 & MoRA & \textit{58.12} & \textit{72.26} & \gearBest[white]{\textit{\textbf{44.40}}} & \textit{80.20} & \textit{45.99} & \gearBest[white]{\textit{\textbf{77.19}}} & \textit{70.01} & \gearBest[white]{\textit{\textbf{77.24}}} & \textit{47.73} & \textit{63.68} \\
 & REAP & 57.40 & 68.17 & 44.20 & 79.92 & 47.61 & 75.51 & 70.48 & 77.21 & \gearBest[white]{\textbf{48.18}} & 63.19 \\
 & MoNE & 62.09 & \gearBest[white]{\textbf{78.47}} & 43.80 & 79.71 & 47.18 & 74.79 & 70.24 & 75.74 & 48.16 & 64.47 \\
 & \gearTableName{} & 62.82 & 76.33 & 43.60 & \gearBest[white]{\textbf{80.52}} & 47.35 & 75.13 & 70.32 & 76.29 & 48.03 & 64.49 \\
\rowcolor{GearFinalFill}\cellcolor{white}\multirow{-7}{*}{\rotatebox[origin=c]{90}{\textit{DeepSeek}}} & \gearMethod{\gearTableNameSquared{}} & \gearBest[GearFinalFill]{\textbf{63.54}} & 76.27 & \gearBest[GearFinalFill]{\textbf{44.40}} & 80.09 & \gearBest[GearFinalFill]{\textbf{47.70}} & 74.33 & 70.72 & 76.56 & 46.94 & \gearBest[GearFinalFill]{\textbf{64.50}} \\
\midrule
 & Original & 67.15 & 85.08 & 48.20 & 83.73 & 68.69 & 89.56 & 76.01 & 83.20 & 74.31 & 75.10 \\
 & REAP & 66.06 & 82.94 & 46.80 & 83.13 & 57.34 & 81.40 & 74.19 & 80.63 & 54.02 & 69.61 \\
 & MoNE & 68.59 & 84.07 & 47.20 & 83.79 & 63.23 & \gearBest[white]{\textbf{86.83}} & 75.45 & \gearBest[white]{\textbf{82.78}} & 66.03 & 73.11 \\
 & \gearTableName{} & \gearBest[white]{\textbf{69.68}} & 83.52 & 46.80 & 84.06 & \gearBest[white]{\textbf{64.42}} & 86.28 & 75.53 & 82.72 & \gearBest[white]{\textbf{66.27}} & 73.25 \\
\rowcolor{GearFinalFill}\cellcolor{white}\multirow{-5}{*}{\rotatebox[origin=c]{90}{\textit{Gemma-4}}} & \gearMethod{\gearTableNameSquared{}} & 68.95 & \gearBest[GearFinalFill]{\textbf{84.37}} & \gearBest[GearFinalFill]{\textbf{49.00}} & \gearBest[GearFinalFill]{\textbf{84.28}} & 63.48 & 86.20 & \gearBest[GearFinalFill]{\textbf{76.48}} & 82.73 & 65.35 & \gearBest[GearFinalFill]{\textbf{73.43}} \\
\midrule
 & Original & 75.09 & 88.23 & 48.20 & 83.46 & 61.86 & 82.66 & 77.51 & 85.87 & 78.93 & 75.76 \\
 & REAP & 69.31 & 87.34 & \gearBest[white]{\textbf{48.60}} & \gearBest[white]{\textbf{83.90}} & 60.67 & 79.42 & \gearBest[white]{\textbf{79.01}} & \gearBest[white]{\textbf{86.06}} & \gearBest[white]{\textbf{74.70}} & 74.33 \\
 & MoNE & 68.95 & 87.74 & 47.20 & 83.62 & 61.77 & 80.39 & 78.69 & 85.87 & 74.53 & 74.31 \\
 & \gearTableName{} & 69.68 & 87.77 & 47.60 & 83.57 & \gearBest[white]{\textbf{61.86}} & \gearBest[white]{\textbf{80.47}} & 78.45 & 85.87 & 74.61 & \gearBest[white]{\textbf{74.43}} \\
\rowcolor{GearFinalFill}\cellcolor{white}\multirow{-5}{*}{\rotatebox[origin=c]{90}{\textit{GLM-4.5}}} & \gearMethod{\gearTableNameSquared{}} & \gearBest[GearFinalFill]{\textbf{70.04}} & \gearBest[GearFinalFill]{\textbf{87.86}} & 47.80 & 83.73 & 61.35 & 79.97 & 78.77 & 85.75 & 74.23 & 74.39 \\
\bottomrule
\end{tabularx}
\caption{Accuracy (\%) on nine zero-shot tasks at 25\% pruning.}
\label{tab:appendix-pertask25}
\end{table*}

%% file: tables/A_paired_ci.tex
\begin{table}[H]
\centering\footnotesize
\setlength{\tabcolsep}{2pt}
\gearTableSetup
\begin{tabularx}{\linewidth}{lYYY}
\toprule
\gearHeaderCell{Model} & \gearHeaderCell{vs.\ MoNE} & \gearHeaderCell{vs.\ REAP} & \gearHeaderCell{vs.\ \gearTableName{}} \\
\midrule
Moonlight & \begin{tabular}[t]{@{}c@{}}$+2.41$\\{\scriptsize[1.76, 3.03]}\end{tabular} & \begin{tabular}[t]{@{}c@{}}$+3.72$\\{\scriptsize[2.82, 4.62]}\end{tabular} & \begin{tabular}[t]{@{}c@{}}$+2.11$\\{\scriptsize[1.48, 2.75]}\end{tabular} \\
Qwen3 & \begin{tabular}[t]{@{}c@{}}$+1.99$\\{\scriptsize[1.46, 2.56]}\end{tabular} & \begin{tabular}[t]{@{}c@{}}$+4.83$\\{\scriptsize[4.02, 5.67]}\end{tabular} & \begin{tabular}[t]{@{}c@{}}$+0.83$\\{\scriptsize[0.34, 1.36]}\end{tabular} \\
DeepSeek & \begin{tabular}[t]{@{}c@{}}$+3.44$\\{\scriptsize[2.65, 4.27]}\end{tabular} & \begin{tabular}[t]{@{}c@{}}$+2.79$\\{\scriptsize[1.84, 3.75]}\end{tabular} & \begin{tabular}[t]{@{}c@{}}$+1.95$\\{\scriptsize[1.22, 2.69]}\end{tabular} \\
Gemma-4 & \begin{tabular}[t]{@{}c@{}}$+2.14$\\{\scriptsize[1.29, 2.96]}\end{tabular} & \begin{tabular}[t]{@{}c@{}}$+5.33$\\{\scriptsize[4.32, 6.35]}\end{tabular} & \begin{tabular}[t]{@{}c@{}}$+1.55$\\{\scriptsize[0.79, 2.32]}\end{tabular} \\
GLM & \begin{tabular}[t]{@{}c@{}}$+0.96$\\{\scriptsize[0.31, 1.64]}\end{tabular} & \begin{tabular}[t]{@{}c@{}}$+0.82$\\{\scriptsize[0.06, 1.56]}\end{tabular} & \begin{tabular}[t]{@{}c@{}}$+1.34$\\{\scriptsize[0.68, 2.02]}\end{tabular} \\
\bottomrule
\end{tabularx}
\caption{Improvement of \gearr{} at 50\% pruning (points), with 95\% paired bootstrap confidence intervals.}
\label{tab:appendix-ci}
\end{table}

%% file: tables/A_reap_mask.tex
\begin{table}[H]
\centering\footnotesize
\setlength{\tabcolsep}{3pt}
\gearTableSetup
\begin{tabularx}{\linewidth}{lYYY}
\toprule
\multicolumn{1}{l}{\gearHeaderCell{\strut Model}} & \multicolumn{1}{c}{\gearHeaderCell{\strut Mean}} & \multicolumn{1}{c}{\gearHeaderCell{\strut\gearTableName{}}} & \multicolumn{1}{c}{\gearHeaderCell{\strut\gearMethod{\gearTableNameSquared{}}}} \\
\midrule
Qwen3 & 64.79 & 65.95 & \textbf{66.62} \\
DeepSeek-V2-Lite & 53.55 & 55.37 & \textbf{57.16} \\
Gemma-4 & 52.12 & 54.56 & \textbf{55.28} \\
\bottomrule
\end{tabularx}
\caption{Nine-task average accuracy (\%) with REAP masks at 50\% pruning. Compensation consistently improves over the mean estimate.}
\label{tab:appendix-reap}
\end{table}

%% file: tables/A_reap_mask_pertask.tex
\begin{table}[t]
\centering\scriptsize
\setlength{\tabcolsep}{0.6pt}
\gearTableSetup
\begin{tabularx}{\linewidth}{lYYYYYYYYY}
\toprule
 & \multicolumn{3}{c}{\gearHeaderCell{Qwen3}} & \multicolumn{3}{c}{\gearHeaderCell{DeepSeek}} & \multicolumn{3}{c}{\gearHeaderCell{Gemma-4}} \\
\cmidrule(lr){2-4} \cmidrule(lr){5-7} \cmidrule(lr){8-10}
\gearHeaderCell{Task} & \gearHeaderCell{Mean} & \gearHeaderCell{\gearTableName{}} & \gearHeaderCell{\gearMethod{\gearTableNameSquared{}}} & \gearHeaderCell{Mean} & \gearHeaderCell{\gearTableName{}} & \gearHeaderCell{\gearMethod{\gearTableNameSquared{}}} & \gearHeaderCell{Mean} & \gearHeaderCell{\gearTableName{}} & \gearHeaderCell{\gearMethod{\gearTableNameSquared{}}} \\
\midrule
RTE & \textbf{78.34} & 75.45 & 75.09 & 53.43 & 55.60 & \textbf{57.04} & 52.35 & \textbf{55.96} & 53.43 \\
BoolQ & \textbf{86.42} & 85.66 & 84.07 & 69.45 & 68.96 & \textbf{72.35} & 69.20 & 68.35 & \textbf{69.76} \\
OBQA & 35.60 & 38.20 & \textbf{40.40} & 36.80 & 36.80 & \textbf{39.40} & 36.00 & \textbf{40.60} & 40.20 \\
PIQA & 76.88 & 77.15 & \textbf{77.58} & 75.19 & 77.04 & \textbf{78.29} & 75.84 & 76.82 & \textbf{78.24} \\
ARC-C & 47.61 & \textbf{49.83} & 49.49 & 34.22 & 37.97 & \textbf{40.10} & 32.00 & 36.86 & \textbf{37.63} \\
ARC-E & 67.97 & 69.95 & \textbf{70.54} & 55.93 & 60.90 & \textbf{64.81} & 55.93 & 58.25 & \textbf{61.95} \\
Wino. & 70.24 & 70.09 & \textbf{70.40} & \textbf{68.03} & 67.88 & 67.09 & \textbf{63.85} & 61.72 & 63.14 \\
Hella. & 66.11 & 67.81 & \textbf{70.39} & 65.92 & 70.00 & \textbf{72.03} & 60.93 & 68.32 & \textbf{69.76} \\
MMLU & 53.90 & 59.39 & \textbf{61.66} & 22.97 & 23.15 & \textbf{23.37} & 22.95 & \textbf{24.20} & 23.44 \\
\midrule
Avg. & 64.79 & 65.95 & \textbf{66.62} & 53.55 & 55.37 & \textbf{57.16} & 52.12 & 54.56 & \textbf{55.28} \\
\bottomrule
\end{tabularx}
\caption{Accuracy (\%) with REAP masks at 50\% pruning.}
\label{tab:appendix-reap-pertask}
\end{table}

%% file: tables/A_robustness.tex
\begin{table}[!ht]
\centering\footnotesize
\setlength{\tabcolsep}{3pt}
\gearTableSetup
\begin{tabularx}{\linewidth}{lYY}
\toprule
\gearHeaderCell{\gearr{}} & \gearHeaderCell{Moonlight} & \gearHeaderCell{DeepSeek} \\
\midrule
Seed 1, rank 32 & 58.51 & 57.15 \\
Seed 2, rank 32 & 59.09 & 56.51 \\
Seed 3, rank 32 & 58.82 & 56.58 \\
Seed 1, rank 16 & 58.44 & 56.27 \\
\midrule
MoNE & 56.68 & 53.71 \\
\bottomrule
\end{tabularx}
\caption{Nine-task average accuracy (\%) of \gearr{} at 50\% pruning across training seeds and ranks.}
\label{tab:appendix-robust}
\end{table}

%% file: tables/A_tau_sweep.tex
\begin{table}[!ht]
\centering\footnotesize
\setlength{\tabcolsep}{3pt}
\gearTableSetup
\begin{tabularx}{\linewidth}{lYYYYY}
\toprule
$\tau$ & 1 & 0.99 & 0.98 & 0.95 & 0.90 \\
\midrule
\gearTableName{} & 51.68 & 56.98 & 56.55 & 56.54 & 56.67 \\
\bottomrule
\end{tabularx}
\caption{Nine-task average accuracy (\%) of GEAR on Moonlight at 50\% pruning under different thresholds; $\tau=1$ disables the gate.}
\label{tab:appendix-tau}
\end{table}

%% file: tables/A_glm_batches.tex
\begin{table}[!ht]
\centering\footnotesize
\setlength{\tabcolsep}{3pt}
\gearTableSetup
\begin{tabularx}{\linewidth}{lYYYY}
\toprule
 & \multicolumn{2}{c}{\gearHeaderCell{25\%}} & \multicolumn{2}{c}{\gearHeaderCell{50\%}} \\
\cmidrule(lr){2-3} \cmidrule(lr){4-5}
\gearHeaderCell{Batch} & \gearHeaderCell{REAP} & \gearHeaderCell{\gearMethod{\gearTableNameSquared{}}} & \gearHeaderCell{REAP} & \gearHeaderCell{\gearMethod{\gearTableNameSquared{}}} \\
\midrule
1 & 74.33 & \textbf{74.39} & 68.81 & \textbf{69.63} \\
2 & 74.16 & \textbf{74.36} & 67.74 & \textbf{69.23} \\
3 & 74.09 & \textbf{74.16} & 68.57 & \textbf{69.85} \\
\bottomrule
\end{tabularx}
\caption{Nine-task average accuracy (\%) of GLM-4.5-Air with three calibration batches. \gearr{} outperforms REAP on every batch.}
\label{tab:appendix-batches}
\end{table}

%% file: tables/A_traj_windows.tex
\begin{table}[H]
\centering\footnotesize
\setlength{\tabcolsep}{3pt}
\gearTableSetup
\begin{tabularx}{\linewidth}{lYYYY}
\toprule
\gearHeaderCell{Model} & \gearHeaderCell{Windows} & \gearHeaderCell{Mean} & \gearHeaderCell{\gearMethod{\gearTableNameSquared{}}} & \gearHeaderCell{Closer} \\
\midrule
Moonlight & 150 & 0.265 & \textbf{0.232} & 111/150 \\
DeepSeek & 150 & 0.222 & \textbf{0.168} & 142/150 \\
GLM & 150 & 0.249 & \textbf{0.239} & 99/150 \\
\bottomrule
\end{tabularx}
\caption{Average hidden-state error at 50\% pruning over token windows, and the number of windows in which \gearr{} is closer to the original model than the mean estimate.}
\label{tab:appendix-traj}
\end{table}